\PassOptionsToPackage{prologue,dvipsnames,table}{xcolor}
\documentclass[sigconf]{acmart}
\usepackage{multirow}
\usepackage[table]{xcolor} 

\usepackage{soul, color, xcolor}
\usepackage{hyperref}  
\usepackage{balance}

\newcommand{\maxvalue}[1]{\cellcolor{red!20}#1} 
\AtBeginDocument{%
  }

\setcopyright{acmlicensed}

\copyrightyear{2025}
\acmYear{2025}
\setcopyright{cc}
\setcctype{by-nc-sa}
\acmConference[MM '25]{Proceedings of the 33rd ACM International Conference on Multimedia}{October 27--31, 2025}{Dublin, Ireland}
\acmBooktitle{Proceedings of the 33rd ACM International Conference on Multimedia (MM '25), October 27--31, 2025, Dublin, Ireland}\acmDOI{10.1145/3746027.3754813}
\acmISBN{979-8-4007-2035-2/2025/10}

\begin{document}

\title{ResearchPulse: Building Method–Experiment Chains through Multi-Document Scientific Inference}


\author{Qi Chen}
\authornote{Both authors contributed equally to this research.}
\authornotemark[0]
\affiliation{%
  \institution{University of Chinese Academy of Sciences}
  \city{Beijing}
  \country{China}
}

\author{Jingxuan Wei}
\authornotemark[1]
\authornote{Corresponding author}
\affiliation{%
  \institution{University of Chinese Academy of Sciences}
  \city{Beijing}
  \country{China}}
\email{weijingxuan20@mails.ucas.edu.cn}

\author{Zhuoya Yao}
\affiliation{%
  \institution{University of Chinese Academy of Sciences}
  \city{Beijing}
  \country{China}}

\author{Haiguang Wang}
\affiliation{%
  \institution{University of Chinese Academy of Sciences}
  \city{Beijing}
  \country{China}}

\author{Gaowei Wu}
\affiliation{%
  \institution{University of Chinese Academy of Sciences}
  \city{Beijing}
  \country{China}}

\author{Bihui Yu}
\affiliation{%
  \institution{University of Chinese Academy of Sciences}
  \city{Beijing}
  \country{China}}

\author{Siyuan Li}
\affiliation{%
  \institution{Zhejiang University}
  \city{Hangzhou}
  \country{China}}

\author{Cheng Tan}
\authornotemark[2]
\affiliation{%
  \institution{Shanghai Artificial Intelligence Laboratory}
  \city{Shanghai}
  \country{China}}
\email{chengtan9907@gmail.com}

\renewcommand{\shortauthors}{Qi Chen et al.}

\begin{abstract}
Understanding how scientific ideas evolve requires more than summarizing individual papers—it demands structured, cross-document reasoning over thematically related research. In this work, we formalize multi-document scientific inference, a new task that extracts and aligns motivation, methodology, and experimental results across related papers to reconstruct research development chains. This task introduces key challenges, including temporally aligning loosely structured methods and standardizing heterogeneous experimental tables. We present \textbf{ResearchPulse}, an agent-based framework that integrates instruction planning, scientific content extraction, and structured visualization. It consists of three coordinated agents: a Plan Agent for task decomposition, a Mmap-Agent that constructs motivation–method mind maps, and a Lchart-Agent that synthesizes experimental line charts. To support this task, we introduce \textbf{ResearchPulse-Bench}, a citation-aware benchmark of annotated paper clusters. Experiments show that our system, despite using 7B-scale agents, consistently outperforms strong baselines like GPT-4o in semantic alignment, structural consistency, and visual fidelity.
The dataset are available in \url{https://huggingface.co/datasets/ResearchPulse/ResearchPulse-Bench}


\end{abstract}




\maketitle

\section{Introduction}
\begin{figure*}[t]
    \centering
    \includegraphics[width=\linewidth]{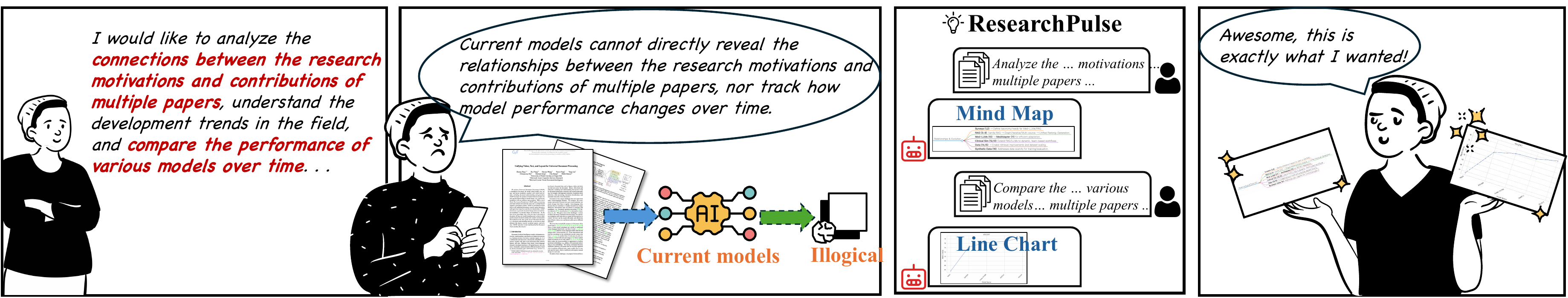}
\caption{ResearchPulse transforms scientific papers into mind maps and line charts for structured research tracking.}    \label{fig:intro}
\end{figure*}
Scientific research is cumulative and comparative. To truly understand how a field evolves, researchers often need to examine multiple thematically related papers together~\cite{zheng2023cvt,tan2025simvpv2,tan2023openstl}, analyze their motivations and methodological innovations, and track how experimental performance changes over time~\cite{caciularu-etal-2023-peek,li2023compressed,zhang2025structvrmaligningmultimodalreasoning}. This type of structured, cross-document analysis is essential for identifying trends, gaps, and breakthroughs in fast-developing domains like artificial intelligence. However, current summarization models~\cite{liu2019text,manakul2021long,zhang2022hegel,zhao2024hierarchical} and automatic survey generators~\cite{liu2023generating,hsu2024chime,wang2024autosurvey,liang2025surveyx} focus either on compressing individual papers or generating high-level overviews, lacking the granularity needed to uncover methodological trajectories and benchmark shifts. Such limitations hinder the ability to track how cutting-edge techniques evolve, converge, or diverge across papers over time. Meanwhile, emerging deep research agents~\cite{yuan2025dolphin,baek2024researchagent,wu2025agentic,tan2025sketchagent,wei2025words,zheng2025deepresearcher} offer promising automation but are either domain-agnostic or lack structural precision. We introduce a new task, multi-document scientific inference, which involves extracting and aligning key research elements—such as motivation, methodology, and experimental results—across thematically related papers to reconstruct structured development chains. As illustrated in Figure~\ref{fig:intro}, this task differs fundamentally from traditional summarization or survey generation, as it requires not only semantic abstraction but also structural parsing and scientific reasoning across documents. This process introduces several core challenges: \textbf{(1) identifying and temporally aligning motivation–method pairs from diverse, loosely structured textual sources}, and \textbf{(2) extracting heterogeneous experimental tables and converting them into unified, interpretable visual trends.} These challenges make multi-document scientific inference a structurally grounded and underexplored problem, distinct from prior efforts in academic summarization or auto-survey pipelines.

To support  this new task, we introduce ResearchPulse-Bench, a benchmark dataset constructed from citation-aware paper clusters curated from \textit{arXiv} and \textit{OpenReview}. Each cluster contains semantically related papers aligned by topic and publication timeline. We annotate core scientific elements—motivations, methods, experimental tables, evaluation metrics, and citation links—using model-assisted extraction techniques, followed by rigorous human validation to ensure structural integrity and factual correctness. This enables high-quality supervision for both Method-Tracking and Experimental-Analysis tasks, supporting model training and standardized evaluation.

Building on this benchmark, we propose ResearchPulse, an end-to-end agent system for structured scientific inference. The system begins with a \textit{Plan Agent} that interprets user instructions and coordinates the workflow. For method-tracking, a \textit{Mmap-Agent} extracts motivation–method pairs from related papers and organizes them into temporally aligned mind maps. For experimental-analysis, a \textit{Lchart-Agent} identifies experimental tables, model names, and evaluation metrics, and synthesizes benchmark trajectories as line charts. Each component operates in a modular yet coordinated manner, enabling ResearchPulse to transform loosely connected papers into coherent, interpretable research chains and support dynamic tracking of scientific progress.

Our contributions are as follows:
\begin{itemize}
    \item We formally define the task of multi-document scientific inference, which focuses on extracting and aligning motivation, method, and experiment elements from thematically related papers to reconstruct structured research development chains.
    \item We introduce \textbf{ResearchPulse-Bench}, a citation-aware benchmark dataset consisting of annotated document clusters and reference outputs for method-tracking and experimental-analysis.
    \item We propose \textbf{ResearchPulse}, a modular agent system that integrates instruction planning, scientific content extraction, and structure-aware visualization into a unified pipeline for tracking methodological evolution and experimental trends in AI research.
\end{itemize}

\section{Related Work}
\subsection{Scientific Document Summarization}  
Scientific document summarization aims to condense long and structurally complex academic texts while preserving core semantic content. Early approaches like TextRank~\cite{mihalcea2004textrank} and LexRank~\cite{erkan2004lexrank} rely on surface-level lexical features and often fail to capture discourse-level dependencies~\cite{koh2022empirical}. Neural and transformer-based models such as BERTSum~\cite{liu2019text} and LoBART~\cite{manakul2021long} introduce deep semantic modeling but remain limited by context length and generation inconsistencies. Later methods like HEGEL~\cite{zhang2022hegel} and HAESum~\cite{zhao2024hierarchical} improve global-local relation modeling through hypergraph and hierarchical attention, though they often depend on noisy tools like LDA or KeyBERT~\cite{vora2024extractive}. Structure-aware models—such as dependency-based discourse parsers~\cite{yoshida2014dependency} and sentence compression with anaphora constraints~\cite{durrett2016learning}—seek better coherence at the cost of increased system complexity. Document expansion~\cite{rao2024single} introduces external context to improve coverage but risks topic drift. Models like BooookScore~\cite{chang2023booookscore} and top-down inference frameworks~\cite{pang2023long} target book-length summarization but still struggle with scientific logic modeling. Overall, while single-document methods evolve in modeling semantics and structure, they often overlook the underlying scientific logic—motivation, method, experiment—and lack mechanisms for linking content across documents.

Multi-document summarization requires integrating redundant or conflicting information from multiple sources. Hybrid frameworks such as Hi-MAP~\cite{hua2024hi}, SKT5SciSumm~\cite{to2024skt5scisumm}, and REFLECT~\cite{song2022improving} combine extraction with generation but often flatten document-specific context. Graph-based systems like CeRA~\cite{gonccalves2023supervising} and cross-document information graphs~\cite{zhang2023enhancing} offer improved factual grounding but mainly focus on entity-event relations. Other methods highlight diversity or specificity—e.g., DisentangleSum~\cite{ma2024disentangling} and DIVERSESUMM~\cite{huang2024embrace}—while large-scale pipelines like GPT-based summarizers~\cite{belem2024single} use recursive generation and clustering to scale. Despite these advances, few approaches address the extraction or alignment of structured scientific reasoning. Evaluation studies confirm this limitation, showing that many benchmarks fail to assess true cross-document synthesis~\cite{ma2024disentangling,huang2024embrace}. These gaps underscore the need for systems that can extract, align, and reason over structured scientific content across thematically related works.

\subsection{Automatic Survey Generation}  
Automatic survey generation seeks to streamline the labor-intensive process of writing literature surveys by leveraging large language models (LLMs). Traditional multi-document summarization methods, designed for small input sets and superficial summaries, fall short for this task. Recent works propose more targeted pipelines. 
For instance, BigSurvey introduces a large-scale dataset and the CAST model, combining sentence classification and sparse-transformer-based generation for structured and abstracted summarization \cite{liu2023generating}. However, such outputs often lack critical synthesis. To enhance organization, CHIME uses hierarchical generation with LLMs to build topic trees and incorporates expert feedback for structural refinement \cite{hsu2024chime}. AutoSurvey adopts a four-stage pipeline—retrieval, outline generation, section-wise writing, and optimization—where multiple LLMs co-generate and critique survey drafts, improving efficiency but still facing depth and grounding issues \cite{wang2024autosurvey}. SurveyX further refines this process through attribute-based preprocessing and information templates, generating attribute forests and novel evaluation metrics for factual consistency and coverage \cite{liang2025surveyx}. Other systems like STORM and ChatCite focus on improving cognitive scaffolding via multi-perspective questioning and human-in-the-loop comparison workflows \cite{shao2024assisting, li2025chatcite}. 
While these approaches mark significant progress in scaling academic synthesis via LLMs, they primarily focus on high-level thematic abstraction, often overlooking fine-grained research trajectories such as methodological evolution or experimental performance shifts. In contrast, our work aims to uncover structural research progressions by aligning methodology and experimental results across thematically linked papers within a specific research direction—thereby enabling more granular  knowledge discovery than generic survey generation pipelines.

\subsection{Automated Deep Research Systems}  
Recent advancements in large language models (LLMs) have enabled deep research systems that go beyond summarization to support iterative scientific inquiry through retrieval, reasoning, and hypothesis generation. For example, DOLPHIN\cite{yuan2025dolphin} introduces a closed-loop framework combining idea generation, experimental validation, and feedback refinement, with features like literature retrieval and error-aware debugging. ResearchAgent\cite{baek2024researchagent} leverages citation graphs and cross-domain knowledge to enhance idea generation, and uses multi-agent LLM-based peer review for evaluating novelty and clarity. Other systems focus on targeted capabilities such as information seeking. DeepSeek-R1\cite{guo2025deepseek} and R1-Searcher\cite{song2025r1} apply reinforcement learning to prompt external search when knowledge gaps arise, using reward-driven prompting and retrieval-aware rollouts to improve factual grounding. Broader agentic reasoning frameworks~\cite{wu2025agentic} coordinate multiple agents—like memory agents, search agents, and code executors—to support multi-hop inference, task decomposition, and synthesis across modalities.
Commercial platforms have also adopted this paradigm. OpenAI\cite{hurst2024gpt} and Gemini\cite{team2024gemini} embed tool use, asynchronous planning, and adaptive web exploration into general-purpose research agents. However, these systems often lack domain-specific precision for structured scientific analysis.

In contrast, our work introduces a focused agent system for structural knowledge discovery within a specific research domain. Rather than generating broad research plans, our system extracts and aligns motivation, methods, and results across related papers—capturing how ideas evolve, techniques recur, and performance shifts over time. This enables fine-grained, structured insights that bridge the gap between general AI agents and the rigorous demands of scientific research.

\section{Method}

\subsection{Task Definition}

We define the objective of \textbf{ResearchPulse} as an agent-based system for scientific document understanding, designed to analyze and organize structural research progress through two complementary tasks: (1) \textit{Method-Tracking}, and (2) \textit{Experimental-Analysis}. Each task enables fine-grained scientific inference over related papers via multi-agent coordination and tool-assisted document processing.

Let $\mathcal{C} = \{D_1, D_2, \dots, D_N\}$ denote a collection of $N$ related scientific documents, typically belonging to the same research thread or citation lineage. Each document $D_i$ contains multiple sections including introduction, methodology, experiments, and references.

\vspace{0.3em}
\noindent\textbf{Method-Tracking Task.}  
Given a document set $\mathcal{C}$ and a user instruction $\mathcal{I}_{\text{method}}$, the goal is to extract for each $D_i$ its core motivation $m_i$ and methodological description $r_i$ from the abstract and introduction. The system then temporally aligns the extracted tuples $\{(m_i, r_i)\}_{i=1}^{N}$ based on the publication timestamp $t_i$ of each $D_i$, and generates a structured representation $\mathcal{M}_{\text{chain}}$ in a hierarchical markdown format. This content is further processed into a mind-map–style visualization.

\vspace{0.3em}
\noindent\textbf{Experimental-Analysis Task.}  
Given the same input document set $\mathcal{C}$ and a user instruction $\mathcal{I}_{\text{exp}}$, the goal is to extract from each $D_i$ its main experimental table $T_i$, associated model names $\mathcal{M}_i$, evaluation metrics $\mathcal{E}_i$, and cited baseline years $\mathcal{Y}_i$. The system organizes the metric values over time and produces a structured summary $\mathcal{E}_{\text{chain}}$ reflecting comparative performance trends. The output is rendered as a line chart via auto-generated Python code.

\subsection{ResearchPulse}

\begin{figure*}[t]
    \centering
    \includegraphics[width=\linewidth]{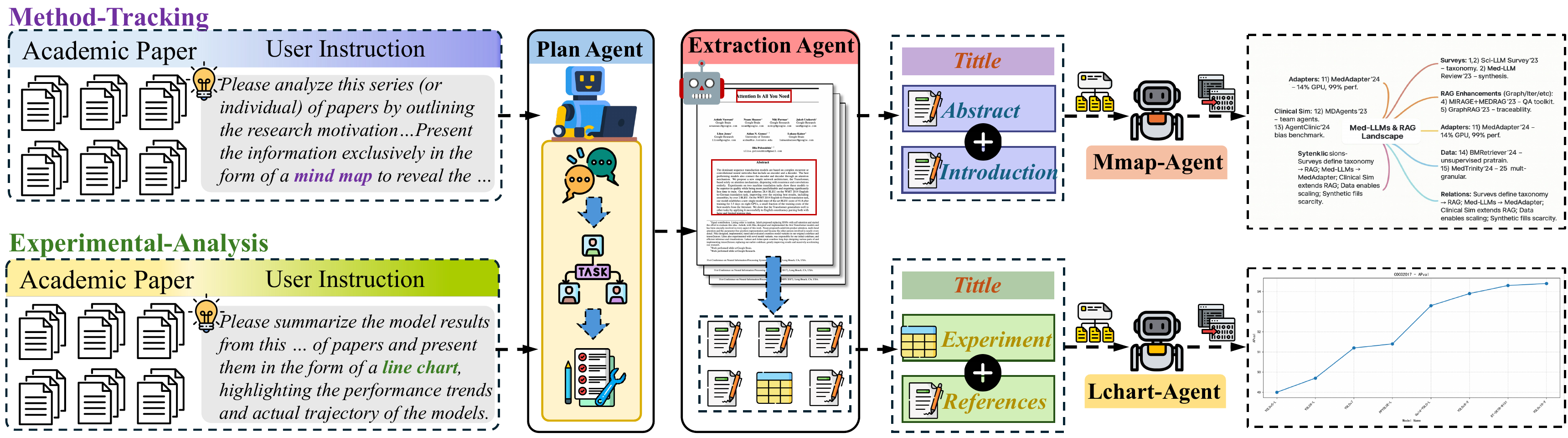}
    \caption{
        The ResearchPulse pipeline, consisting of three main agents: Plan Agent, Mmap-Agent, and Lchart-Agent.
    }
    \label{fig:pipeline}
\end{figure*}

ResearchPulse is implemented as a modular, instruction-driven agent system designed to perform structured scientific inference across related documents. Given a document set $\mathcal{C} = \{D_1, D_2, ..., D_N\}$ and a user-issued instruction $\mathcal{I}$, the system decomposes the task into two coordinated subtasks—Method Tracking and Experimental Analysis—each handled by a specialized sub-agent. The overall pipeline is illustrated in Figure~\ref{fig:pipeline}.

\subsubsection{Plan Agent}

The \textit{Plan Agent} acts as the central controller of ResearchPulse. It receives user instruction $\mathcal{I}$ and identifies the corresponding intent category: $\mathcal{I}_{\text{method}}$ for method-tracking or $\mathcal{I}_{\text{exp}}$ for experimental comparison. Based on this classification, it dynamically dispatches subtasks to the appropriate downstream agent (Mmap-Agent or Lchart-Agent).

To support end-to-end automation, the Plan Agent invokes an Extraction module, which parses each document $D_i$ in $\mathcal{C}$ into its structured components—including abstract, introduction, experiment tables, and references—producing normalized content $X_i$:
\begin{equation}
    X_i = \text{Extract}(D_i), \quad \forall i \in \{1, \dots, N\}.
\end{equation}

The agent then routes each $X_i$ with the appropriate instruction $\mathcal{I}_{\text{method}}$ or $\mathcal{I}_{\text{exp}}$ to the respective downstream processing agent. This decouples instruction planning from content processing, enabling modular coordination across diverse scientific tasks.

\subsubsection{Mmap-Agent}

The Mmap-Agent is responsible for identifying the research motivation and methodology for each paper in the corpus and constructing a time-aligned representation of scientific evolution. Specifically, for each $X_i$, the agent extracts a tuple $(m_i, r_i)$ representing the motivation and method, respectively. These are typically sourced from the abstract and introduction sections.

Formally, this agent performs:
\begin{equation}
    (m_i, r_i) = \mathcal{F}_{\text{Mmap}}(X_i, \mathcal{I}_{\text{method}}),
\end{equation}
where $\mathcal{F}_{\text{Mmap}}$ is a fine-tuned LLM designed for structural scientific information extraction. The output for all documents is then temporally sorted by publication timestamp $t_i$ to produce a markdown-formatted research chain:
\begin{equation}
    \mathcal{M}_{\text{chain}} = \text{Sort}_{t_i} \left( \{(m_i, r_i)\}_{i=1}^{N} \right),
\end{equation}
which is subsequently rendered into a mind map visualization.

To supervise this extraction process, we optimize the negative log-likelihood of the predicted motivation-method tuple sequence:
\begin{equation}
    \mathcal{L}_{\text{Mmap}} = - \sum_{i=1}^{N} \log P((m_i, r_i) \mid X_i, \mathcal{I}_{\text{method}}),
\end{equation}
where $P(\cdot)$ denotes the conditional probability predicted by the fine-tuned LLM. The training data includes manually annotated pairs of motivations and methods across representative paper series. The loss encourages the agent to capture salient reasoning structures and align them chronologically for subsequent visualization.

\subsubsection{Lchart-Agent}

The Lchart-Agent is designed to extract experimental results from each paper and generate visual representations of comparative performance. For each document $X_i$, the agent identifies its main experimental table $T_i$, model names $\mathcal{M}_i$, evaluation metrics $\mathcal{E}_i$, and baseline publication years $\mathcal{Y}_i$ linked via citation resolution.

Formally, the agent conducts:
\begin{equation}
    (T_i, \mathcal{M}_i, \mathcal{E}_i, \mathcal{Y}_i) = \mathcal{F}_{\text{Lchart}}(X_i, \mathcal{I}_{\text{exp}}),
\end{equation}
where $\mathcal{F}_{\text{Lchart}}$ is a separately fine-tuned agent optimized for tabular extraction and metric normalization. The extracted results are aligned across time and datasets to construct a structured trend record:
\begin{equation}
    \mathcal{E}_{\text{chain}} = \text{Align} \left( \{(T_i, \mathcal{M}_i, \mathcal{E}_i, \mathcal{Y}_i)\}_{i=1}^N \right),
\end{equation}
which is visualized as a line chart through Python code automatically generated by the agent:
\begin{equation}
    \text{Chart} = \mathcal{V}(\mathcal{E}_{\text{chain}}).
\end{equation}

The Lchart-Agent is trained to extract structured experimental content with high precision. Its objective is to maximize the correctness and completeness of extracted tuples $(T_i, \mathcal{M}_i, \mathcal{E}_i, \mathcal{Y}_i)$ under supervision. We define the training loss as:
\begin{equation}
    \mathcal{L}_{\text{Lchart}} = - \sum_{i=1}^{N} \log P(T_i, \mathcal{M}_i, \mathcal{E}_i, \mathcal{Y}_i \mid X_i, \mathcal{I}_{\text{exp}}),
\end{equation}
where the probability is computed over multi-field outputs. The agent is fine-tuned using a curated dataset of aligned tables and metric annotations, encouraging accurate parsing of experimental results and consistent temporal alignment across documents.


\begin{figure*}[t]
    \centering
    \includegraphics[width=\linewidth]{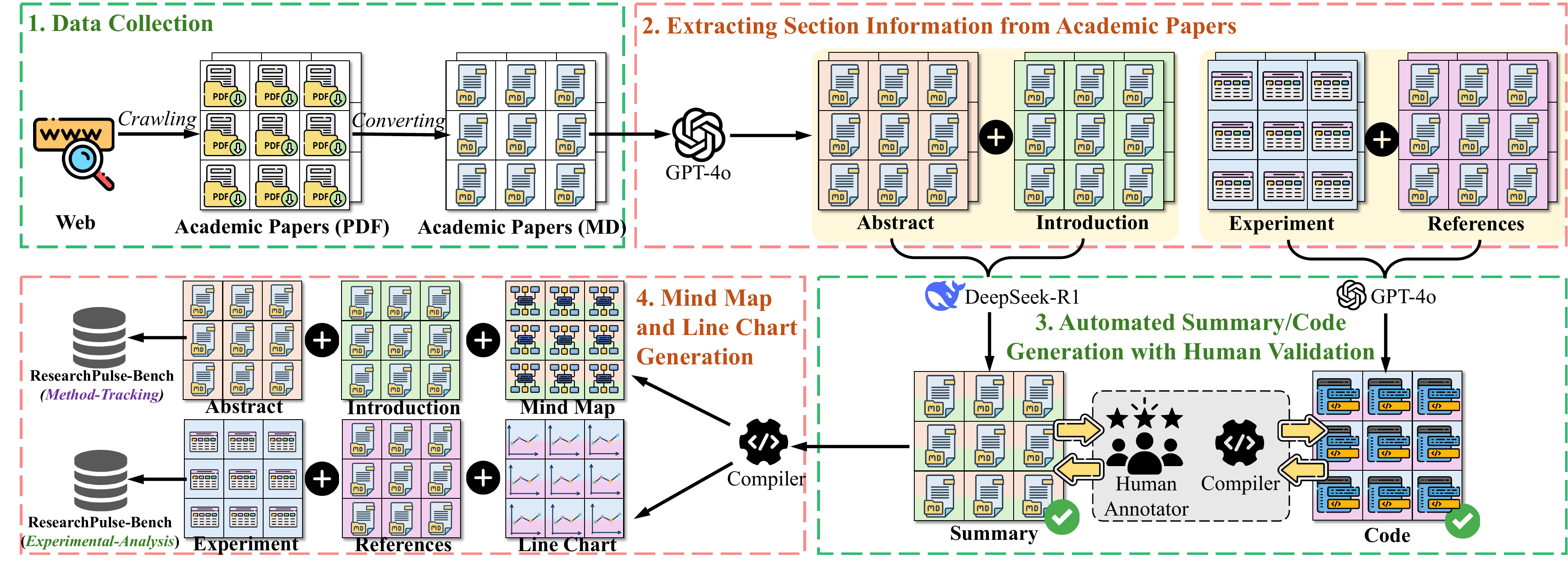}
    \caption{The construction pipeline of ResearchPulse-Bench.}
    \label{fig:data_pipeline}
\end{figure*}
\section{ResearchPulse-Bench}

We present \textbf{ResearchPulse-Bench}, a benchmark designed to support multi-document scientific inference tasks including method-tracking and experimental comparison. Unlike traditional summarization datasets, ResearchPulse-Bench is constructed from real-world citation networks and explicitly captures the structure of research progressions. Each instance contains a series of related scientific papers organized by semantic and temporal proximity, annotated with motivation, methodology, and experimental evidence. The pipeline includes data collection, document parsing, information extraction, and visualization, as shown in Figure~\ref{fig:data_pipeline}.

\paragraph{Data Collection.}
We begin by retrieving a set of seed papers from Google Scholar based on AI-related keywords (e.g., ``Machine Learning'', ``Computer Vision'', ``Natural Language Processing''), citation counts, and publication timeframes (past five years). To ensure data openness, we restrict our collection to papers that are publicly accessible via platforms such as \textbf{arXiv} and \textbf{OpenReview}, complying with their open-access terms. Each selected paper's metadata—including title, authorship, publication date, and citation statistics—is extracted using public APIs or permitted crawling mechanisms. We then construct citation graphs by retrieving forward and backward references, enabling us to group semantically and temporally related papers into document clusters for downstream processing.

\paragraph{Data Processing.}
The collected papers are first parsed using PDF-to-markdown conversion tools to extract structured content. Section segmentation is performed to isolate the abstract, introduction, experiment, and reference sections. For semantic clustering, we encode each abstract with Sentence-BERT to generate vector representations, then apply K-Means to group papers by topic proximity. Each resulting cluster is labeled using high-frequency terms and reviewed manually. Within each cluster, we use DeepSeek-R1 or GPT-4o~\cite{gpt4o} to extract fine-grained elements including motivations, methods, and experimental results. These outputs are stored in markdown format and further rendered as mind map diagrams and temporal trend charts to facilitate structured inspection of research motivations, methodologies, and experimental trajectories. This processed data is subsequently used for visual generation, as illustrated in Figure~\ref{fig:data_pipeline}.

\paragraph{Human Inspection.}
To ensure dataset reliability, we conduct a multi-stage human inspection. Annotators review the semantic clustering results to verify topical coherence and identify misassigned papers. For method-related content, we check whether motivations and methodologies are logically and chronologically consistent. For experiment-based summaries, we verify if extracted tables match their originals, whether citation years are correctly resolved, and whether metric values are consistent with ground truth. All visual outputs—mind maps and line charts—are reviewed for structural completeness and semantic fidelity. Annotator feedback is used to adjust clustering granularity and refine LLM prompts, forming a closed-loop quality assurance process.

\begin{table}[t]
  \centering
  \caption{Token statistics for the Method-Tracking and Experimental-Analysis tasks.}
  \resizebox{\linewidth}{!}{
    \begin{tabular}{lrrrr}
    \toprule
    \multirow{2}[2]{*}{\textbf{Statistic}} & \multicolumn{2}{c}{\textbf{Method-Tracking}} & \multicolumn{2}{c}{\textbf{Experimental-Analysis}} \\
          & \multicolumn{1}{l}{train} & \multicolumn{1}{l}{test} & \multicolumn{1}{l}{train} & \multicolumn{1}{l}{test} \\
    \midrule
    \multicolumn{5}{l}{{\textbf{Total Samples}}} \\
    Sample Count & 1958  & 491   & 1550  & 320 \\
    \midrule
    \multicolumn{5}{l}{{\textbf{Query Length (tokens)}}} \\
    Minimum & 55    & 78    & 301   & 3026 \\
    Maximum & 22326 & 24031 & 46877 & 32971 \\
    Average & 1161.16 & 1210.78 & 14402.73 & 16545.45 \\
    \midrule
    \multicolumn{5}{l}{{\textbf{Answer Length (tokens)}}} \\
    Minimum & 63    & 141   & 331   & 407 \\
    Maximum & 1722  & 2036  & 3081  & 2020 \\
    Average & 344.54 & 355.6 & 885.98 & 928 \\
    \bottomrule
    \end{tabular}%
    }
  \label{tab:token_stats}%
\end{table}%
\subsection{Data Analysis}
\paragraph{Dataset Scale and Token Statistics.}
ResearchPulse-Bench contains 100 citation-aware document clusters, each curated for either the Method-Tracking or Experimental-Analysis task. Table~\ref{tab:token_stats} summarizes the number of samples as well as the input/output token lengths. Method-Tracking queries are relatively concise, averaging around 1.1K tokens per input and 350 tokens per output. Experimental-Analysis queries are significantly longer (averaging over 14K tokens) and require structured extraction of tabular data, with outputs around 900 tokens. This range reflects the dual challenge of focused scientific summarization and long-context reasoning.

\begin{figure*}[t]
    \centering
    \includegraphics[width=\linewidth]{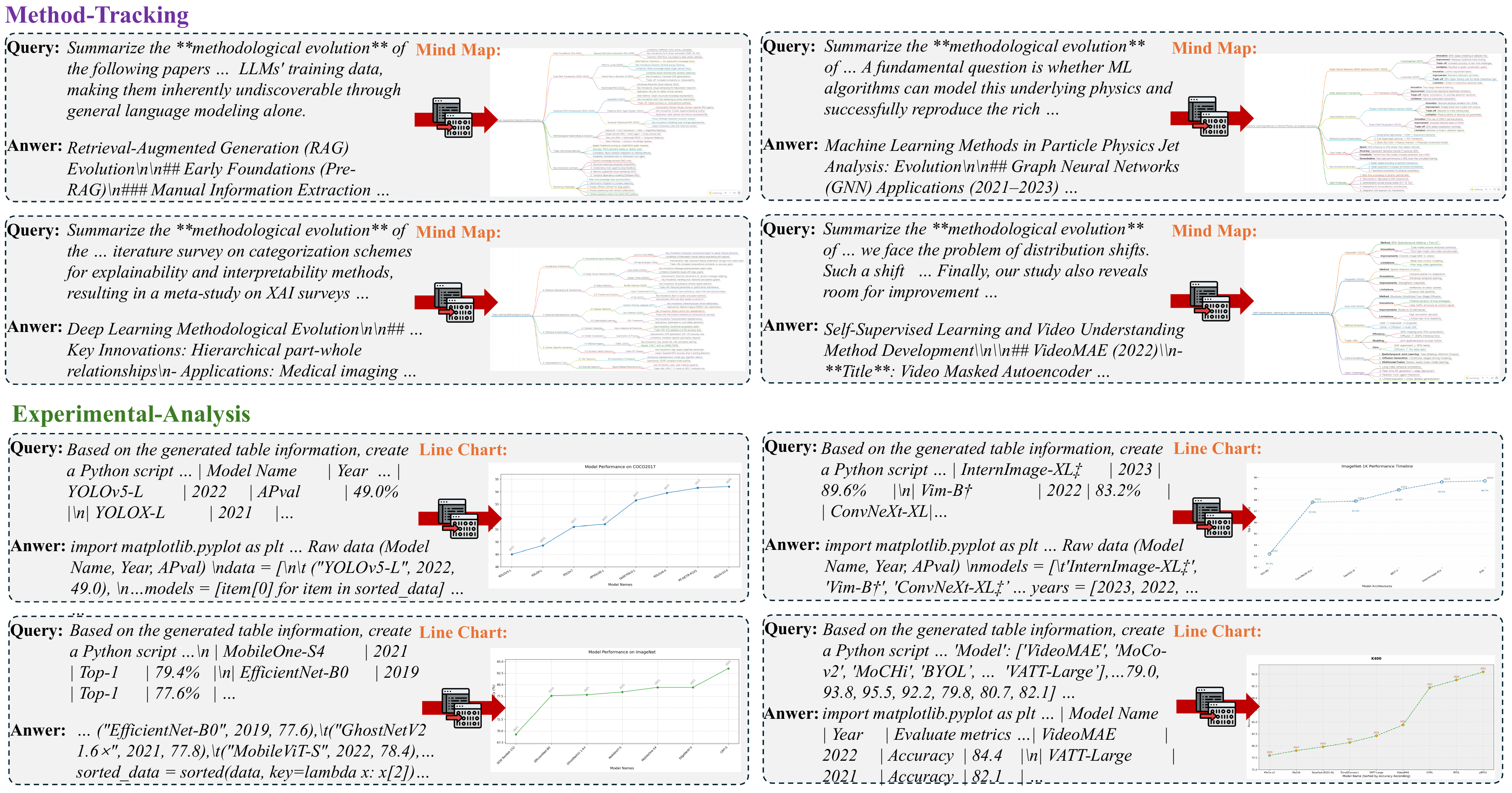}
    \caption{Representative examples from Method-Tracking (top) and Experimental-Analysis (bottom) tasks.}
    \label{fig:data_examples}
\end{figure*}
\paragraph{Cluster Distribution and Examples.}
\begin{table}[t]
  \centering
  \setlength{\tabcolsep}{5mm}
  \caption{Distribution of paper clusters.}
    \begin{tabular}{lrrr}
    \toprule
    \textbf{Series Count} & \textbf{Train} & \textbf{Test} & \textbf{Total} \\
    \midrule
    Total & 80    & 20    & 100 \\
    Minimum Papers & 6     & 8     & 14 \\
    Maximum Papers & 33    & 31    & 38 \\
    Average Papers & 19.19 & 19.9  & 24.93 \\
    \bottomrule
    \end{tabular}
  \label{tab:series_stats}
\end{table}

Table~\ref{tab:series_stats} shows the distribution of document clusters across train and test splits. Each cluster forms a temporally ordered and semantically coherent series of papers. On average, clusters contain around 20 documents, with the largest containing over 30. This structure supports fine-grained modeling of research trajectories over time. Representative examples for both tasks are shown in Figure~\ref{fig:data_examples}, including a mind map from method-tracking and a line chart from experimental-analysis. These outputs highlight the benchmark's diversity and its support for downstream tasks such as trend forecasting and methodological comparison.

\subsection{Evaluation Metrics}

We employ a comprehensive set of evaluation metrics tailored to the multimodal outputs of ResearchPulse. For the method-tracking task, which produces structured textual summaries of motivations and methodologies, we use BERTScore (Precision, Recall, F1) and METEOR to assess semantic similarity and lexical overlap with references. To capture qualitative aspects such as fluency, factuality, and coherence, we additionally report GPT-Score—a human-aligned rating between 0 and 1 generated by GPT-4o along five dimensions: fluency, relevance, accuracy, creativity, and overall quality.

In the experimental-analysis task, the system outputs visual charts derived from extracted tables. To evaluate these generated figures, we adopt standard image quality metrics including Inception Score (IS), Fréchet Inception Distance (FID), Kernel Inception Distance (KID), and CLIP-FID for perceptual and distributional alignment. We further assess structural fidelity using SSIM, MS-SSIM, PSNR, LPIPS, and CMMD.

When the system generates Python code to render visualizations, we evaluate functional correctness using the pass@1 metric, which reports the percentage of code generations that execute successfully without runtime errors. Together, these metrics comprehensively evaluate the system’s outputs across semantics, visual quality, and functional reliability.

\section{Experiment}

\paragraph{Setup.} ResearchPulse performs method-tracking and experimental-analysis through four core agents: Plan Agent, Mmap-Agent, and Lchart-Agent. The Plan Agent, utilizing Qwen-72B\cite{qwen25}, is responsible for effectively distinguishing and interpreting the user's instructions. The Mmap-Agent is based on Qwen2.5-7B~\cite{Qwen2.5-7B-Instruct}, while the Lchart-Agent leverages Qwen2.5-Coder-7B~\cite{Qwen2.5-Coder-7B-Instruct}. Both agents were fine-tuned for four epochs on a 4×80GB A100 GPU setup to optimize performance for their respective tasks. 

\begin{table*}[h]
    \centering
    \caption{(a) Main results: Performance comparison of Mmap-Agent with state-of-the-art models. The best result is highlighted in bold. (b) Ablation study: Impact of different module configurations on Mmap-Agent performance.}
    \label{tab:method-tracking}
    {\renewcommand\baselinestretch{1.2}\selectfont
    \resizebox{\linewidth}{!}{
    \begin{tabular}{l|c|c|c|c|c|c|c|c|c|c}
    \hline
        \multirow{2}{*}{\textbf{Model}} & \multirow{2}{*}{\textbf{Size}} & \multirow{2}{*}{\textbf{METEOR↑}} & \multicolumn{3}{c|}{\textbf{BERTScore}} & \multicolumn{5}{c}{\textbf{GPT-4o Score}} \\ \cline{4-11}
        ~ & ~ & & \textbf{P↑} & \textbf{R↑} & \textbf{F1↑} & \textbf{Fluency↑} & \textbf{Relevance↑} & \textbf{Accuracy↑} & \textbf{Creativity↑} & \textbf{Quality↑} \\ \midrule
        \multicolumn{11}{c}{(a) Main results}                                       \\ \midrule
        Qwen2.5-7B-Instruct & 7B & 42.60 & 88.80 & 86.23 & 87.49 & 92.10 & 87.14 & 81.57 & 78.72 & 85.10 \\ 
        InternLM3-8B-instruct & 8B & 46.14 & 87.46 & 87.25 & 87.35 & 89.45 & 85.21 & 80.34 & 78.43 & 83.46 \\ 
        Llama-3.1-8B-Instruct & 8B & 39.47 & 88.64 & 85.80 & 87.19 & 90.44 & 84.52 & 77.57 & 74.96 & 82.00 \\ 
        CodeLlama & 7B & 17.49 & 81.94 & 81.79 & 81.85 & 80.50 & 80.10 & 73.07 & 73.69 & 81.59 \\ 
        Qwen2.5-Coder-7B-Instruct & 7B & 41.03 & 88.78 & 85.74 & 87.22 & 91.68 & 86.62 & 80.71 & 78.65 & 84.57 \\ 
        GPT-4o & - & 42.66 & 84.90 & 86.06 & 85.48 & 88.32 & 82.46 & 78.27 & 75.88 & 83.42 \\ 
        Claude-3.7-Sonnet & - & 42.73 & 88.43 & 85.34 & 86.85 & 90.63 & 85.99 & 80.32 & 78.40 & 84.01 \\
        Gemini-1.5 Pro & - & 43.25 & 88.75 & 83.42 & 85.99 & 90.26 & 85.48 & 79.95 & 77.14 & 83.32 \\ 
        Mmap-Agent & 7B & \maxvalue{\textbf{46.14}} & \maxvalue{\textbf{90.90}} & \maxvalue{\textbf{89.89}} & \maxvalue{\textbf{90.39}} & \maxvalue{\textbf{92.16}} & \maxvalue{\textbf{87.62}} & \maxvalue{\textbf{82.29}} & \maxvalue{\textbf{79.54}} & \maxvalue{\textbf{85.59}} \\ \midrule
        \multicolumn{11}{c}{(b) Ablation study}                                       \\ \midrule
        w/o GPT-4o & 7B & 41.78 & 87.23 & 87.93 & 86.91 & 91.29 & 86.49 & 81.45 & 73.88 & 83.98 \\
        w/o Compiler & 7B & 44.13 & 88.19 & 84.67 & 83.49 & 90.18 & 83.73 & 80.18 & 76.11 & 81.69 \\ 
        w/o GPT-4o \& Compiler & 7B & 42.23 & 86.11 & 83.99 & 82.83 & 90.05 & 82.99 & 79.64 & 75.32 & 82.71 \\ \hline
    \end{tabular}}}
\end{table*}

\paragraph{Model.} In the method-tracking and experimental-analysis tasks, Mmap-Agent and Lchart-Agent are compared with several state-of-the-art models, including both open-source and closed-source approaches. The open-source models include Qwen2.5-7B-Instruct~\cite{Qwen2.5-7B-Instruct}, InternLM3-8B-Instruct~\cite{InternLM3-8B-instruct}, Llama-3.1-8B-Instruct~\cite{Llama-3.1-8B-Instruct}, CodeLlama-7B-hf~\cite{codellama}, and Qwen2.5-Coder-7B-Instruct~\cite{Qwen2.5-Coder-7B-Instruct}. Closed-source models include GPT-4o~\cite{gpt4o}, Claude-3.7-Sonnet~\cite{Claude-3.7-Sonnet}, and Gemini-1.5 Pro~\cite{gemini}. These models were selected based on their advanced performance in similar tasks and serve as key benchmarks for evaluating Mmap-Agent and Lchart-Agent.

\subsection{Method-Tracking}

\paragraph{Main Results} Mmap-Agent achieves superior performance across multiple evaluation metrics, as demonstrated in Table~\ref{tab:method-tracking}. It leads in BERTScore F1 (90.39) and METEOR (46.14), showcasing its exceptional ability to preserve semantic accuracy and generate fluent, contextually appropriate text. In GPT-4o-based evaluations, Mmap-Agent surpasses other models with an overall score, excelling in fluency (92.16), relevance (87.62), accuracy (82.29), creativity (79.54) and quality (85.59). While baseline models like InternLM3-8B-Instruct and Claude-3.7-Sonnet demonstrate strong performance in specific areas, such as METEOR or fluency, they fall short in providing balanced performance across all dimensions. InternLM3, for instance, achieves the highest METEOR score (46.14), but its BERTScore F1 (87.35) and GPT-4o results are lower, indicating weaker semantic alignment and generation quality. Similarly, Claude-3.7-Sonnet excels in fluency (90.63) and relevance (85.99), but underperforms in accuracy (80.32) and creativity (78.40). In contrast, Mmap-Agent’s well-rounded performance across fluency, relevance, accuracy, and creativity highlights its robustness in both linguistic fluency and semantic integrity. These results validate Mmap-Agent’s strength in method-level reasoning and generation tasks, where it consistently outperforms other models by generating coherent, precise, and diverse text.

\paragraph{Ablation Study} To evaluate the impact of key components on Mmap-Agent’s performance, we conducted ablation studies by systematically removing critical modules. As illustrated in Table~\ref{tab:method-tracking}, the removal of the GPT-4o module resulted in notable declines in fluency and creativity, with BERTScore F1 dropping to 86.91. When the Compiler module was excluded, there was a marked reduction in precision and relevance, further lowering BERTScore F1 to 83.49. Removing both modules simultaneously caused an additional performance drop, with BERTScore F1 falling to 82.83. These results emphasize the importance of both modules in maintaining Mmap-Agent’s overall performance, particularly in ensuring high fluency, precision, and relevance in generated text.

\begin{table*}[h]
    \centering
    \caption{(a) Main results: Performance comparison of Lchart-Agent with state-of-the-art models. The best result is highlighted in bold. (b) Ablation study: Impact of different module configurations on Lchart-Agent performance.}
    \label{tab:experimental-analysis}
    {\renewcommand\baselinestretch{1.2}\selectfont
    \resizebox{\linewidth}{!}{
    \begin{tabular}{l|c|c|c|c|c|c|c|c|c|c|c}
    \hline
        \textbf{Model} & \textbf{Size} & \textbf{Pass@1↑} & \textbf{IS↑} & \textbf{FID↓} & \textbf{KID↓} & \textbf{CLIP-FID↓} & \textbf{LPIPS↓} & \textbf{CMMD↓} & \textbf{SSIM↑} & \textbf{PSNR↑} & \textbf{MS-SSIM↑} \\  \midrule
        \multicolumn{12}{c}{(a) Main results}                                       \\ \midrule
        Qwen2.5-7B-Instruct & 7B & 65.00 & 0.18 & 12.73 & 4.10 & 2.09 & 24.35 & 9.85 & 29.13 & 7.93 & 26.51 \\ 
        InternLM3-8B-instruct & 8B & 68.13 & 0.57 & 29.91 & 2.08 & 4.42 & 10.39 & 15.22 & 22.97 & 5.14 & 17.91 \\ 
        Llama-3.1-8B-Instruct & 8B & 58.75 & 0.24 & 36.39 & 2.14 & 1.64 & 13.47 & 6.70 & 24.71 & 6.55 & 15.17 \\ 
        CodeLlama-7B-hf & 7B & 55.00 & 0.09 & 30.16 & 5.33 & 0.23 & 30.32 & 10.63 & 20.83 & 9.18 & 10.58 \\ 
        Qwen2.5-Coder-7B-Instruct & 7B & 73.13 & 0.77 & 10.69 & 2.15 & 5.50 & 16.71 & 12.02 & 32.33 & 6.37 & 21.18 \\ 
        GPT-4o & - & 96.25 & 2.40 & 7.15 & \maxvalue{\textbf{1.14}} & 1.18 & 9.31 & 4.09 & 53.76 & 12.14 & 23.70 \\ 
        Claude-3.7-Sonnet & - & 91.56 & 2.28 & 8.56 & 3.65 & \maxvalue{\textbf{0.08}} & 9.81 & 7.24 & 52.25 & 11.90 & 36.34 \\ 
        Gemini-1.5 Pro & - & 90.31 & 1.10 & 8.68 & 4.33 & 5.75 & 10.40 & 5.47 & 42.33 & 9.57 & 23.25 \\ 
        Lchart-Agent & 7B & \maxvalue{\textbf{97.50}} & \maxvalue{\textbf{2.65}} & \maxvalue{\textbf{6.73}} & 1.17 & 1.10 & \maxvalue{\textbf{8.49}} & \maxvalue{\textbf{4.06}} & \maxvalue{\textbf{55.56}} & \maxvalue{\textbf{12.49}} & \maxvalue{\textbf{36.55}} \\ \midrule
        \multicolumn{12}{c}{(b) Ablation study}                                       \\ \midrule
        w/o GPT-4o & 7B & 97.50 & 2.32 & 7.13 & 1.25 & 1.22 & 9.13 & 5.08 & 52.77 & 10.76 & 33.52 \\ 
        w/o Compiler & 7B & 90.63 & 2.26 & 8.02 & 1.37 & 1.14 & 9.22 & 6.16 & 51.67 & 11.53 & 26.85 \\ 
        w/o GPT-4o \& Compiler & 7B & 90.00 & 1.98 & 8.33 & 1.56 & 1.31 & 9.38 & 7.32 & 48.79 & 9.84 & 22.91 \\ \hline
    \end{tabular}}}
\end{table*}

\subsection{Experimental-Analysis}

\paragraph{Main Results} As shown in Table~\ref{tab:experimental-analysis}, Lchart-Agent outperforms most models across key metrics, achieving a Pass@1 score of 97.50, surpassing GPT-4o (96.25) and Claude-3.7-Sonnet (91.56). It also achieves the lowest FID of 6.73, demonstrating superior image quality. Additionally, Lchart-Agent excels in perceptual metrics like LPIPS (8.49), CMMD (4.06), and SSIM (55.56), indicating high visual and semantic consistency. However, other models show strengths in specific areas: Claude-3.7-Sonnet leads in CLIP-FID with 0.08, suggesting better semantic alignment, while GPT-4o outperforms in KID with 1.14, indicating sharper images. Despite these areas of strength for other models, Lchart-Agent maintains a competitive edge overall, consistently excelling in generating high-quality, semantically consistent images across multiple dimensions.

\paragraph{Ablation Study} As shown in Table~\ref{tab:experimental-analysis}, removing the GPT-4o module resulted in a decrease in IS to 2.32 and an increase in FID to 7.13, indicating the module's crucial role in maintaining high-quality image generation. Similarly, excluding the Compiler module led to a reduction in performance, with Pass@1 dropping to 90.63 and FID rising to 8.02. Notably, the Compiler module also plays a key role in ensuring code executability, as evidenced by its direct impact on the Pass@1 score. When both modules were removed together, Pass@1 further declined to 90.00 and FID increased to 8.33, reinforcing the importance of these components in optimizing the model’s overall performance. 

\subsection{Error Analysis}

\begin{figure}[!h]
    \centering
    \includegraphics[width=\linewidth]{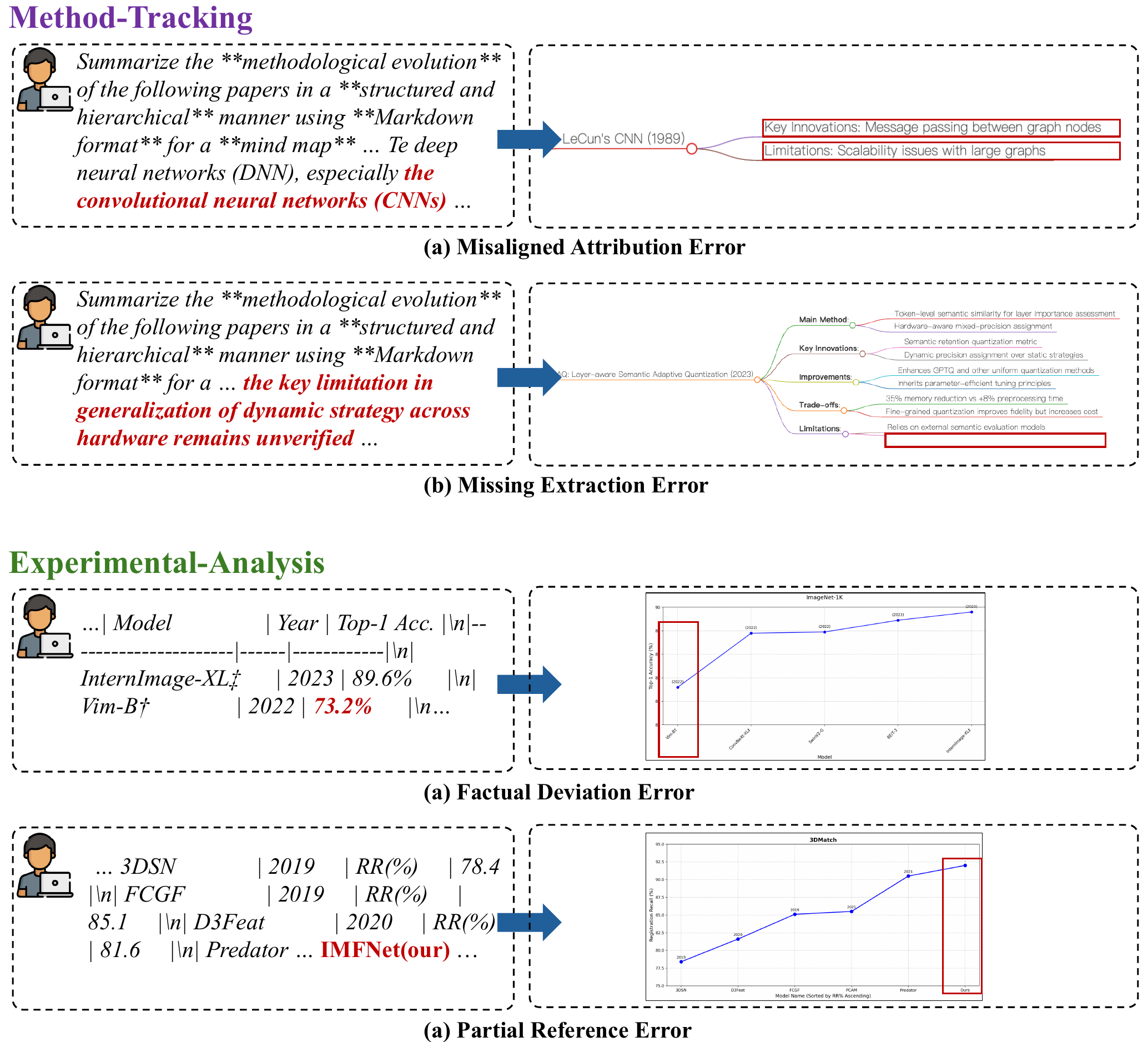}
    \caption{The error examples of Mmap-Agent and Lchart-Agent.}
    \label{fig:error}
\end{figure}

We conduct an in-depth analysis of typical failure cases across the Method-Tracking and Experimental-Analysis tasks, as illustrated in Figure~\ref{fig:error}. In the Method-Tracking task, a prominent issue is the misaligned attribution error, where the model assigns key innovations to incorrect sources—for example, attributing “message passing between graph nodes” to LeCun’s CNN (1989), despite it being a defining feature of later graph-based neural networks such as GCNs or GATs. This indicates a misunderstanding of foundational methodological distinctions. Another frequent error is missing extraction, where the model overlooks essential components of the described method. In the case of layer-aware semantic adaptive quantization, the system fails to capture innovations such as dynamic precision assignment and semantic-aware evaluation metrics, resulting in incomplete or superficial summarization. For the Experimental-Analysis task, we observe factual deviation errors, such as incorrect reporting of Top-1 accuracy or misinterpretation of data trends from visualized results, which can undermine the credibility of the analysis. Furthermore, partial reference errors are evident when models mention results (e.g., IMFNet) without sufficient contextual grounding or comparative clarity. These observations underscore the necessity of more robust reasoning and alignment mechanisms to ensure faithful extraction and accurate interpretation of complex scientific content.

\section{Conclusion}

We present \textbf{ResearchPulse}, a modular agent system designed to perform multi-document scientific inference by extracting, aligning, and visualizing key research elements across thematically related papers. The system supports two complementary tasks: \textit{Method-Tracking}, which constructs motivation–method chains rendered as mind maps, and \textit{Experimental-Analysis}, which synthesizes benchmark trajectories from structured experimental results. To enable high-quality supervision, we introduce \textbf{ResearchPulse-Bench}, a citation-aware benchmark comprising 100 annotated paper clusters curated from arXiv and OpenReview. Experiments show that our agents, despite operating at 7B scale, achieve state-of-the-art performance in semantic alignment and multimodal output quality, outperforming both open- and closed-source baselines. Our work offers a new paradigm for tracking research evolution and structuring scientific knowledge at scale.


\bibliographystyle{unsrt}

\appendix




\end{document}